\documentclass{bmvc2k}

\usepackage{graphicx}
\usepackage{multirow}
\usepackage{amsmath}
\usepackage{amsthm}
\usepackage{bm}
\usepackage{booktabs}

\usepackage{algorithmic}
\usepackage[ruled,linesnumbered]{algorithm2e}

\title{Center-wise Local Image Mixture  \\ For Contrastive Representation Learning}

\addauthor{Hao Li}{hao_li_96@163.com}{1}
\addauthor{Xiaopeng Zhang}{zxphistory@gmail.com}{2}
\addauthor{Hongkai Xiong}{xionghongkai@sjtu.edu.cn}{1}

\addinstitution{
 Institute of M.I.N \\
 Shanghai Jiao Tong University\\
 Shanghai, China
}
\addinstitution{
 Huawei, Inc.\\
 Shanghai, China
}

\runninghead{HAO ET AL.}{Center-wise Local Image Mixture}

\begin{document}

\maketitle

\begin{abstract}

Contrastive learning based on instance discrimination trains model to discriminate different transformations of the anchor sample from other samples, which does not consider the semantic similarity
among samples.
This paper proposes a new kind of contrastive learning method, named CLIM, which uses positives
from other samples in the dataset. This is achieved by searching local similar samples of the anchor, and selecting samples that are closer to the corresponding cluster center, which we denote as center-wise local image selection. The selected samples are instantiated via an data mixture strategy, which performs as a smoothing regularization. As a result, CLIM encourages both local similarity and global aggregation in a robust way, which we find is beneficial for feature representation. Besides, we introduce \emph{multi-resolution} augmentation, which enables the representation to be scale invariant. We reach $75.5\%$ top-1 accuracy with linear evaluation over ResNet-50, and $59.3\%$ top-1 accuracy when fine-tuned with only $1\%$ labels.
\end{abstract}

\section{Introduction}
\label{sec:intro}

 Recently, self-supervised learning has attracted more attention due to its free of human labels. In self-supervised learning, the network aims at exploring the intrinsic distributions of images via a series of predefined pretext tasks \cite{gidaris2018unsupervised,noroozi2016unsupervised}.
Among them,  contrastive learning pulls closer together the positive pairs composed by different transformations of the same image, \emph{e.g.,} cropping, color distortion, \emph{etc.}, and has been demonstrated to be able to generate features that are comparable with those produced by supervised pretraining \cite{chen2020big}. However, contrasting two images that are \emph{de facto} similar in semantic space is not optimal for general representations. It is intuitive to pull semantically similar images for better transferability. DeepCluster \cite{caron2018deep} and Local Aggregation \cite{zhuang2019local} relax the extreme instance discrimination task via discriminating groups of images instead of an individual image. However, due to the lack of labels, it is inevitable that the positive pairs contain noisy samples, which limits the performance.

In this paper, we target at expanding instance discrimination by exploring local similarities and facilitate transitivity among images. Towards this goal, we need to solve two issues: i) how to select similar images as positive pairs, and ii) how to incorporate these positive pairs, which inevitably contain noisy assignments, into contrastive learning. This paper proposes a new kind of contrastive learning method, named \emph{Center-wise Local Image Mixture}, to tackle the above two issues in a robust way. 
CLIM consists of two core elements, \emph{i.e.,} a center-wise positive sample selection, as well as a data mixing operation. 
For center-wise sample selection, we search for nearest neighbors of an image, and only retain similar samples that are closer to the corresponding cluster center. As a result, an image is pulled towards the corresponding center without breaking the local similarity.
Considering that the selected positive samples inevitably contain noisy samples, we apply data mixing augmentation as a regularization strategy to avoid predictions with high confidence on selected positive samples. In this way, similar samples are pulled together in a smoother and robust way, which we find is beneficial for general representation.


Furthermore, we propose \emph{multi-resolution} augmentation, which aims at contrasting the same image (patch) at different resolutions explicitly, to enable the representation to be scale invariant. We argue that although previous operations such as crop and resize introduce multi-resolution implicitly, they do not compare the same patch at different resolutions directly. As comparisons, multi-resolution incorporates scale invariance into contrastive learning, and significantly boosts the performance even based on a strong baseline.


We evaluate the feature representation on several self-supervised learning benchmarks. On ImageNet linear evaluation protocol, we achieve $75.5\%$ top-1 accuracy with a standard ResNet-50, and achieve $59.3\%$ top-1 accuracy when finetuned with only $1\%$ labels. We also validate its transferring ability on several downstream tasks, and consistently outperform the fully supervised counterparts.

\begin{figure}[t]
  \begin{center}
        \includegraphics[width=0.8\linewidth]{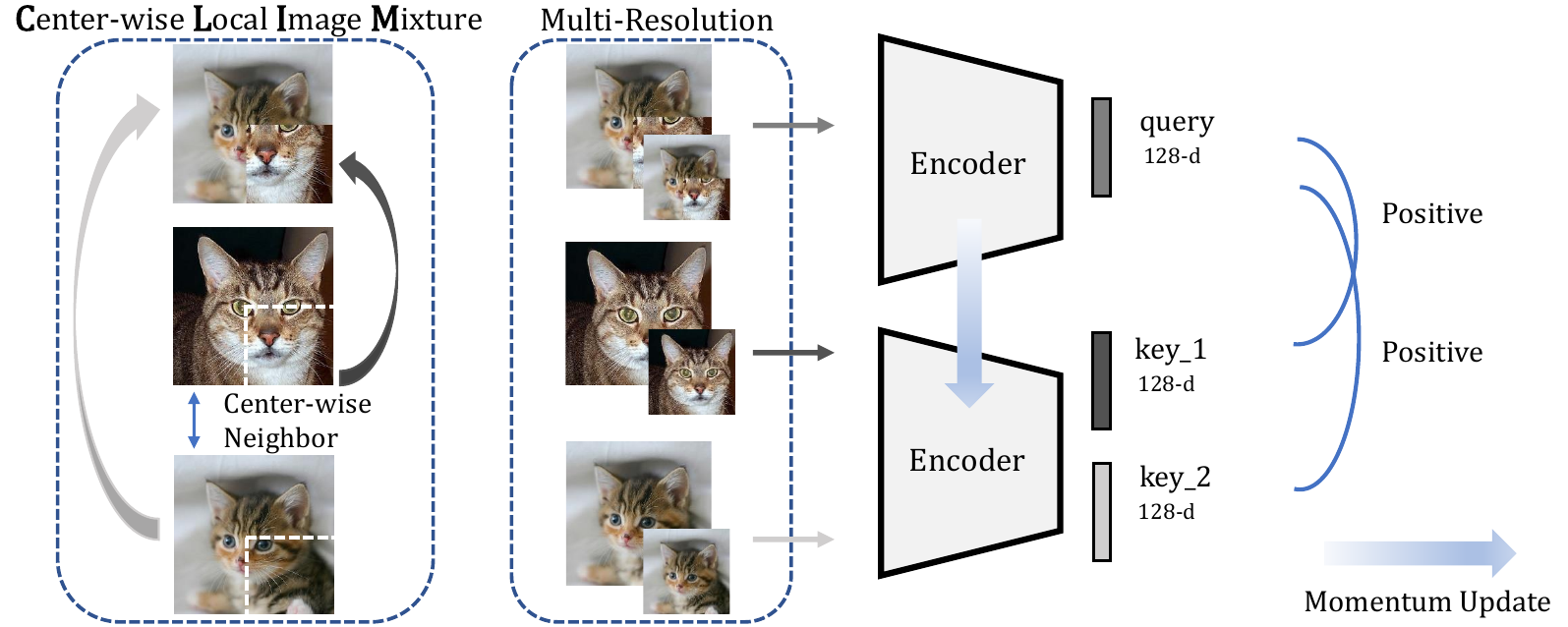}
  \end{center}
  \vspace{-0.5cm}
     \caption{An illustration of the proposed \textbf{CLIM} and \textbf{Multi-resolution} data augmentations.}
  \label{main_stru}
\end{figure}

\section{Related Work}

\textbf{Unsupervised Representation Learning.} Unsupervised learning aims at exploring the intrinsic distribution of data samples via constructing a series of pretext tasks without human labels. These pretext tasks take many forms and vary in utilizing different properties of images. Among them, one family of methods takes advantage of the spatial properties of images, typical pretext tasks include predicting the relative spatial positions of patches \cite{noroozi2016unsupervised}, or inferring the missing parts of images by colorization \cite{zhang2016colorful}, or rotation prediction \cite{gidaris2018unsupervised}. Recent progress in self-supervised learning mainly benefits from instance discrimination, which regards each image (and augmentations of itself) as one class for contrastive learning. The motivation behind these works is the InfoMax principle, which aims at maximizing mutual information \cite{tian2019contrastive,wu2018unsupervised} across different augmentations of the same image \cite{he2020momentum,chen2020simple, tian2019contrastive}.

Contrastive learning does not consider the relationship between different samples. Some related works \cite{han2020self, miech2020end} propose to use different mining strategies to select positive samples.  CoCLR \cite{han2020self} proposes to use rgb flow and optical flow to select positive samples alternately and interactively. In \cite{miech2020end}, the authors propose MIL-NCE to learn a joint embedding space where semantically related videos and texts are close and far away otherwise. Both these two methods use other modalities to facilitate the samples selection. Our proposed method, CLIM selects the positive samples from the nearest neighbors with a direction provided by cluster center,  which helps to pull together semantically similar images.

\noindent \textbf{Data Augmentation.} Instance discrimination relies on data augmentations, \emph{e.g.,} random cropping, color jittering, horizontal flipping, to define a large set of vicinities for each image. As has been demonstrated in \cite{chen2020simple,tian2020makes}, the effectiveness of instance discrimination methods strongly relies on the type of augmentations, hoping that the network holds invariance in the local vicinities of each sample.  

Mixing different images is widely used as a data augmentation to help alleviate overfitting in training deep networks. In particular, Mixup \cite{zhang2017mixup} combines two samples linearly on pixel level, where the target of the synthetic image was a linear combination of one-hot labels. Following Mixup, there are a few variants \cite{yun2019cutmix,li2020attribute}. Cutmix \cite{yun2019cutmix} cuts
out a patch from image, pastes it on another image, and mix their labels according to the area proportion. Attribute Mix \cite{li2020attribute} mixes the attentive regions to generate new samples.

In contrastive learning, Un-mix \cite{shen2020mix} uses data mixing augmentation to expand data space, which is totally different from CLIM. Without the restriction of labels, the improvements Un-mix brings are limited.  CLIM uses data mixing method to regularise the relationship between positive sample pairs. The data mixing operations are only performed on positive sample pairs. 

\section{Method}
\label{Method}
In this section, we start by reviewing contrastive learning for unsupervised representation learning. Then we elaborate our proposed CLIM method, which targets at pulling similar samples via center-wise similar sample selection, followed by a data mixing regularization. We also present multi-resolution augmentation that further improves the performance.

\subsection{Contrastive Learning}
Contrastive learning targets at training an encoder to map positive pairs to similar representations while pushing away the negative samples in the embedding space. Given unlabeled training set $\bm{X}=\left \{ x_{1},x_{2},...,x_{n} \right \}$. Instance-wise contrastive learning aims to learn an encoder $f_{q}$ that maps the samples $\bm{X}$ to embedding space $\bm{V}=\left \{ v_{1},v_{2},...,v_{n} \right \}$ by optimizing a contrastive loss. Take the Noise Contrastive Estimator (NCE) \cite{oord2018representation} as an example, the contrastive loss is defined as:

\begin{small}

\begin{equation}
  \label{InfoNCE}
  \mathcal{L}_{nce}(x_{i},{x}'_{i}) =  -\log\frac{\exp(\frac{f_{q}(x_{i})\cdot f_{k}({x}'_{i})}{\tau})}{\exp(\frac{f_{q}(x_{i})\cdot f_{k}({x}'_{i})}{\tau})+\sum\limits_{j=1}^{N}\exp(\frac{f_{q}(x_{i})\cdot f_{k}({x}'_{j})}{\tau})}
\end{equation}
\end{small}
where $\tau$ is the temperature parameter, and ${x}'_{i}$ and ${x}'_{j}$ denote the positive and negative samples of ${x}_{i}$, respectively. The encoder $f_{k}$ can be shared \cite{chen2020simple,caron2020unsupervised} or momentum update of encoder $f_{q}$~\cite{he2020momentum}. $N$ denotes the number of negative samples sampled.

\subsection{CLIM: Center-wise Local Image Mixture}
\label{Section::CLIM}

In contrastive learning, each sample as well as its transformations are treated as a separate class, while all other samples are regarded as negative examples. In principle, semantically similar samples should be endowed with similar feature representation in the embedding space, while current contrastive methods does not consider the semantic similarities among different samples. To solve this issue, we propose a new contrastive learning method, termed as CLIM, pulls together samples that are semantically similar in an efficient and robust way. The proposed CLIM consists of two elements, \emph{i.e.,} center-wise local similar sample selection, and a data mixing regularization, which would be described in details in the following.

\begin{figure*}[!t]
  \begin{center}
        \includegraphics[width=1\linewidth]{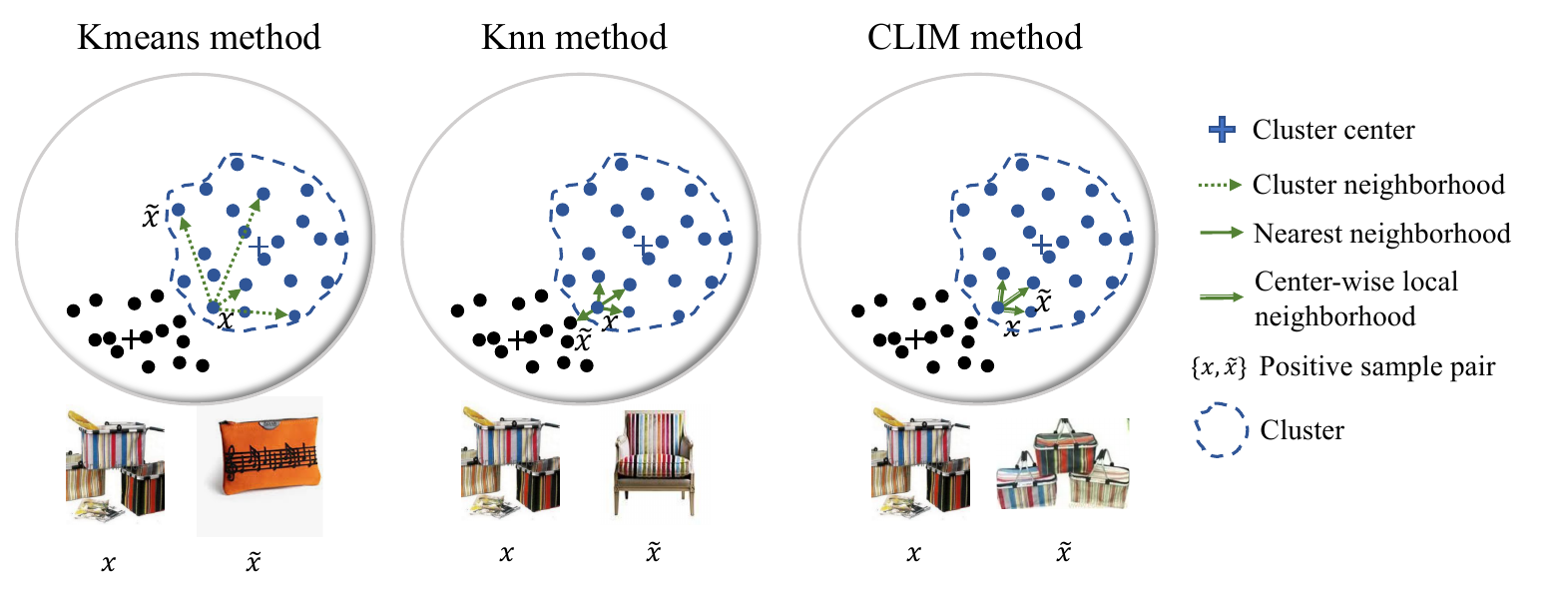}
  \end{center}
  \vspace{-0.1cm}
     \caption{Comparison of three positive sample selection strategies, \emph{i.e.,} K-means, Knn, and the proposed center-wise local sample selection.}
  \label{three_select}
\end{figure*}

\textbf{Center-wise Local Positive Sample Selection.}

To facilitate transitivity among samples, we propose a positive sample selection strategy that considers both local similarity and global aggregation to expand the neighborhood space of current anchor sample. This is achieved by searching similar samples within a cluster that the anchor sample belongs to, and only retaining samples that are closer to the corresponding cluster center. We denote it as center-wise local selection as these samples are picked out towards the cluster center among the local neighborhood of an image. In this way, similar samples are progressively pulled to the predefined cluster centers, while do not break the local similarity.

Specifically, given a set of unlabeled images $\bm{X}\!=\!\left \{ x_{1},x_{2},...,x_{n} \right \}$ and the corresponding embedding $\bm{V}=\left \{ v_{1},v_{2},...,v_{n} \right \}$ with encoder $f_{\theta}$, where $v_i=f_{\theta}(x_i)$. We cluster the representations $\bm{V}$ using a standard K-means algorithm, and obtain $m$ centers $\mathbf{C}= \left \{ c_{1},c_{2},...,c_{m} \right \}$. Given an anchor $x_i$ with its assigned cluster $c(x_i) \in \bm{C}$, denote the sample set that belongs to $c(x_i)$ as $\bm{\Omega}_1 = \{x|c(x)=c(x_i)\}$. We search the $k$ nearest neighbors of $x_i$ over the entire space with L2 distance, obtaining sample set $\bm{\Omega}_2 = \{x_{i1},...,x_{ik}\}$. The positive samples are selected based on the following rule:

\begin{equation}
  \label{positive_sample}
  \bm{\Omega}_{p} = \{x|d(f_{\theta}(x),v_{c(x_i)})\leq d(f_{\theta}(x_i), v_{c(x_i)}), x \in \bm{\Omega}_{1} \cap \bm{\Omega}_{2}\}
\end{equation}

where $d(\cdot,\cdot)$ denotes the L2 distance of two samples, and $v_{c(x_i)}$ denotes the feature representation of the corresponding cluster center, respectively. In this way, the samples are aggregated towards the predefined clusters, and meanwhile maintaining the local similarity.

Our method combines the advantages of cluster and nearest neighbor methods. An illustration comparing the three methods is shown in Fig. ~\ref{three_select}. Cluster-based method regards all samples that belong to the same center as positive pairs, which breaks the local similarity among samples especially when the anchor is around the boundary. While nearest neighbor-based method independently pulling samples of an anchor, and does not encourage the well-clustered goal. As a result, the embedding space is not highly concentrated among multiple similar anchors. As comparisons, by center-wise sample selection, similar samples are progressively pulled to the predefined center as well as considering the local similarity. In Fig.~\ref{three_select}, some visualizations of positive sample pairs are provided.  We used MoCo pre-trained model to infer on the validation set and do K-means on the output features. Almost $48\%$ samples in this cluster are from the same category. The remaining $52\%$ of the samples, especially those on the cluster boundaries, have no similarities either in layout or category information. We also use KNN to search Top-K neighbors of every sample on the output features space. It was found that that the nearest neighborhoods of the anchor sample have at least one of these two characteristics: sharing the same category or sharing the similar layout and color information with the anchor sample. CLIM integrates the advantages of KNN and K-means, and selects positive samples sharing similar category, layout and color information with the anchor sample.   In the experimental section, we would demonstrate its advantages over simply cluster and knn in feature representation.

\textbf{Data Mixing Regularization.}
Once we obtain the positive samples of an anchor, one direct way is to treat these samples similar as the augmented ones for contrastive learning. However, similarity computation in high dimensional space is complex, which will bring the issue of uncertainty and trust-worthiness. To solve this issue, we make use of data mixture strategy, which aims at mixing patches from two different images as augmented samples for contrasting. Data mixing is widely used in supervised learning as
data augmentation.
Here data mixing is used as a regularization strategy to avoid high confidence for the selected positive samples.

We mix selected positive sample with the anchor sample, and mixed samples obtained can be used to constitute positive pairs $(x_{mix},\tilde{x}_{i})$ and $(x_{mix},x_{i})$.
In this way, the selected sample $\tilde{x}_{i}$ and anchor $x_{i}$ can be pulled together in a soft manner. Specifically, we conduct data mixing as Cutmix~\cite{yun2019cutmix}, which can be described as:

\begin{equation}
  \label{mix_eq}
  \begin{aligned}
      x_{mix} &= \mathbf{M}\odot x_{i}+(\bm{1}-\mathbf{M})\odot \tilde{x}_{i}
  \end{aligned}
\end{equation}

where $\mathbf{M} \in \{0,1\}^{W \times H}$ denotes a binary mask indicating the mixed rectangle region of an image, \emph{i.e.,} where to cutout the region in $x_i$ and replaced with a randomly selected patch from $\tilde{x}_i$, and $W,H$ denotes the wide and height of an image, respectively. $\mathbf{1}$ is a binary mask filled with ones, and $\odot$ is the element-wise multiplication operation. For mask $\mathbf{M}$ generation, we follow the setting in ~\cite{yun2019cutmix}. For the mixed sample $x_{mix}$, the positive sample can be either $x_i$ or $\tilde{x}_{i}$, and we reformulate the contrastive learning as combing two NCE loss:

\begin{equation}
  \label{mixNCE}
  \mathcal{L}_{mix}(x_{i},\tilde{x}_{i}) = \lambda \cdot \mathcal{L}_{nce}(x_{mix},{x}_{i}) + (1 - \lambda) \cdot \mathcal{L}_{nce}(x_{mix},\tilde{x}_{i})
\end{equation}
where the combination ratio $\lambda$ is sampled from beta distribution Beta$(\alpha,\alpha)$ with parameter $\alpha$. The advantages are twofold: first, mixed samples help to expand the neighborhood space of current anchor sample for better representation; second, minimizing the two terms simultaneously can help to maximize the mutual information between $x_{i}$ and $\tilde{x}_{i}$ in a soft manner and perform as smoothing regularization on the prediction for selected positive samples.

\subsection{Multi-resolution Data Augmentation}
\label{Section::MR}
Data augmentation plays a key role in contrastive learning, crop augmentation is one of the most effective way \cite{chen2020simple}. In a typical crop augmentation, a sample $x$ with size $H \times W$ is randomly cropped with ratio $\sigma$, and resized to $K_{train} \times K_{train}$ as augmented samples, where $K_{train} \times K_{train}$ denotes the input resolution for model training. Hence the scaling factor w.r.t. sample $x$ can be described as:
\begin{equation}
  \label{mr}
  s=\frac{1}{\sigma}\cdot \frac{K_{train}}{\sqrt{H \times W}}.
\end{equation}

For crop augmentation, the parameter $K_{train}$ is fixed, and the crop ratio $\sigma$ is randomly selected among positive pairs. As a result, different crop augmentations usually contain different contents, which can be regarded as modeling occlusion invariance to some extent, where each crop sees one view of an image. In this section, we propose a simple but effective data augmentation strategy, named multi-resolution augmentation, which enables the representation to be scale invariant of an example. The highlight is that it is better for contrasting positive pairs with the same content but different resolutions. Specifically, for each positive we keep the  crop ratio $\sigma$ fixed, and adjust $K_{train}$ to different resolutions for contrastive loss. An illustration is shown in Fig.~\ref{main_stru} .Using multi-resolution, the objective function can be generalized as:

\begin{equation}
  \label{mr_obj}
  \mathcal{L}_{mr} = \sum_{r,r'\in \{ r_{1},...,r_{n}\}}\mathcal{L}_{mix}(x^{r}_{i},\tilde{x}^{r'}_{i}),
\end{equation}
where $\{r_{1},..., r_{n}\}$ indicates the resolution set. In this way, the encoder would be encouraged to discriminate the positive samples with different resolutions from a series of negative keys, which will maximize the mutual information between inputs with different resolutions and discard redundant information brought by resolutions.

\textbf{Comparing with Multi-crop Augmentation.} There exist recent works that aim at improving crop augmentations, including multi-crop~\cite{caron2020unsupervised} and jigsaw-crop~\cite{misra2020self}. However, both methods target at reducing crop ratio $\sigma$ in Eq.\ref{mr} and resolution $K_{train}$ simultaneously to bridge different parts of an object, and do not explicitly model scale invariance. As comparisons, our proposed multi-resolution strategy fixes the crop ratio to explicitly model scale invariance. In the experimental section, we would validate the discrepancy of the two augmentation strategies.

\section{Experimental Results}
In this section, we assess our pretrained feature representation on several unsupervised benchmarks. We evaluate it on ImageNet under linear evaluation and semi-supervised settings. Then we transfer the learned features to different downstream tasks. We also analyze the performance of our representation with detailed ablation studies. For brief expression, except for the ablation study, we denote our method as CLIM, which includes two kinds of data augmentations.



\begin{figure}[t]
\begin{minipage}[t!]{\textwidth}
 \begin{minipage}[t]{0.55\textwidth}
  \centering
     \makeatletter\def\@captype{table}\renewcommand\arraystretch{1.25}\setlength{\abovecaptionskip}{0.3cm}\setlength{\belowcaptionskip}{0.2cm}\makeatother\caption{Top-1 accuracies under linear evaluation on ImageNet, using ResNet-50 as encoder}\label{linear-table}
        \begin{tabular}{lr}
            \hline
            \multicolumn{1}{l}{ Method} & \multicolumn{1}{r}{Accuracy (\%)}
             \\ \hline
            Supervised &  76.5  \\
            \hline
            Colorization~\citep{zhang2016colorful}       &    39.6   \\
            Jigsaw~\citep{noroozi2016unsupervised}       &    45.7   \\
            NPID~\citep{wu2018unsupervised}              &    54.0   \\
            LA~\citep{zhuang2019local}                   &    58.8   \\
            MoCo~\citep{he2020momentum}                  &    60.6   \\
            SeLa~\citep{asano2020self}                   &    61.5   \\
            PIRL~\citep{misra2020self}                   &    63.6   \\
            CPCv2~\citep{henaff2019data}                 &    63.8   \\
            PCL~\citep{li2020prototypical}               &    65.9   \\
            SimCLR~\citep{chen2020simple}                &    70.0   \\
            MoCo v2~\citep{chen2020improved}              &    71.1   \\
            SimCLRv2~\citep{chen2020big}                 &    71.7   \\
            InfoMin~\citep{tian2020makes}                &    73.0   \\
            BYOL~\citep{grill2020bootstrap}              &    74.3   \\
            SwAV~\citep{caron2020unsupervised}           &    75.3   \\
            \hline
            CLIM                                &    \textbf{75.5}  \\

            \hline
        \end{tabular}
  \end{minipage}
  \hspace{0.1cm}
  \begin{minipage}[t]{0.42\textwidth}
  \begin{minipage}[t]{1\textwidth}
   \centering
        \makeatletter\def\@captype{table}\renewcommand\arraystretch{1.1}\setlength{\abovecaptionskip}{0.25cm}\renewcommand\tabcolsep{2pt}\setlength{\belowcaptionskip}{0cm}\makeatother\caption{Semi-supervised learning with few shot ImageNet labels, using ResNet-50 as encoder (averaged by 5 trials)}\label{semi-table}
        \begin{tabular}{lcccc}
            \hline
            \multirow{2}{*}{ Method}                                                  &   \multicolumn{4}{c}{ Top-1 / Top-5}     \\ \cline{2-5}
                      &  \multicolumn{2}{c}{ 1$\%$ labels}  & \multicolumn{2}{c}{10$\%$ labels}   \\
            \hline
            Supervised                    &  25.4 &  48.4    & 56.4 & 56.4       \\
            \hline
            PIRL                          &  30.7 &  57.2    & 60.4 & 83.8       \\
            SimCLR                        &  48.3 &  75.5    & 65.6 & 87.8       \\
            MoCo v2     &  52.4 &  78.4    & 65.3 & 86.6       \\
            BYOL                          &  53.2 &  78.4    & 68.8 & 89.0        \\
            SwAV                          &  53.9 &  78.5    & \textbf{70.2} & \textbf{89.9}   \\
            SimCLRv2                      &  57.9 &  \textbf{82.5}    & 68.4 & 89.2         \\
            \hline
            CLIM                           &  \textbf{59.3} &  81.6 & 70.0 & 89.3   \\
            \hline
        \end{tabular}
   \end{minipage}
  \begin{minipage}[b]{1\textwidth}
   \centering
        \makeatletter\def\@captype{table}\renewcommand\arraystretch{1.1}\setlength{\abovecaptionskip}{0.25cm}\setlength{\belowcaptionskip}{0cm}\makeatother\caption{Transfer learning on VOC object detection (averaged by 5 trials)}\label{voc-table}
        \begin{tabular}{lcc}
            \hline
            \multirow{2}{*}{ Method} &   \multicolumn{2}{c}{Accuracy ($\%$)}    \\ \cline{2-3}
                         &  \multicolumn{1}{c}{ AP$_{50}$}  & \multicolumn{1}{c}{ AP$_{75}$}   \\
            \hline
            Supervised                                       &  81.4     & 58.8                  \\
            \hline
            MoCo v2      &  82.5     & 64.0            \\
            SwAV                                          &  82.6     & -            \\
            \hline
            CLIM                                             &  \textbf{82.8}  & \textbf{64.5}           \\
            \hline
        \end{tabular}
   \end{minipage}
   \end{minipage}
\end{minipage}
\vspace{0.1cm}
\end{figure}


\subsection{Linear evaluation on ImageNet} \label{linear}
The feature representation is trained based on ImageNet 2012 \citep{russakovsky2015imagenet}, using a standard ResNet-50 structure as backbone. We follow the setting in MoCo v2 \citep{chen2020improved}, and the training details are listed in Appendix. We first evaluate our features by training a linear classifier on top of the frozen representation, following a common protocol in \citep{he2020momentum,tian2019contrastive}. For linear classifier, the learning rate is initialized as 30 and decayed by 0.1 after 60, 80 epochs, respectively. Table.\ref{linear-table} shows the top-1 accuracies with center crop evaluation. Our method achieves an accuracy of $75.5\%$, surpassing MoCo v2 baseline ($71.1\%$) by $4.4\%$, and nearly approaching the supervised learning baseline ($76.5\%$).

\subsection{Semi-supervised training on ImageNet}
We also evaluate our method by fine-tuning the pretrained model with a small subset of labels, following the semi-supervised settings in \citep{grill2020bootstrap, kornblith2019better, chen2020simple, caron2020unsupervised}. For fair comparisons, we use the same fixed $1\%$ and $10\%$ splits of training data as in \citep{chen2020simple}, and fine-tune all layers using SGD optimizer with momentum of 0.9, and learning rate of 0.0001 for backbone, 10 for the newly initialized fc layer. The fine-tune epochs is set as 60, and the learning rate is decayed by 0.1 after every 20 epochs. During training, only random cropping and flipping data augmentations are used for fair comparison. The results are reported in Table.\ref{semi-table}. CLIM achieves $59.3\%$ top-1 accuracy with only $1\%$ labels, and $70.0\%$  with $10\%$ labels. The performance gains are larger with $1\%$ labels, \emph{e.g.,} $6.1\%$ higher than BYOL, and $5.4\%$ better than SwAV, which demonstrates that the proposed feature representation is mainly suitable for extremely few shot learning. Note that SimCLR v2 makes use of other tricks like more MLP layers for better performance, while our method simply adds one fc layer, and still achieves better performance under both settings.

\subsection{Downstream tasks}

\begin{table}[t]
\caption{Transfer learning on COCO detection and instance segmentation (5 trials)}
\label{coco-table}
\setlength{\abovecaptionskip}{0.1cm}
\setlength{\belowcaptionskip}{0cm}
\renewcommand\arraystretch{1.1}
\renewcommand\tabcolsep{2pt}
\begin{center}
\begin{tabular}{l|ccc|ccc|ccc|ccc}
\hline
\multicolumn{1}{c|}{\multirow{3}{*}{Method}} & \multicolumn{6}{c|}{Mask R-CNN,R50-FPN,Det}                                                       & \multicolumn{6}{c}{Mask R-CNN,R50-FPN,InsSeg}                                                    \\ \cline{2-13}
\multicolumn{1}{c|}{}                         & \multicolumn{3}{c|}{1$\times$ schedule}               & \multicolumn{3}{c|}{2$\times$ schedule}               & \multicolumn{3}{c|}{1$\times$ schedule}               & \multicolumn{3}{c}{2$\times$ schedule}               \\ \cline{2-13}
\multicolumn{1}{c|}{}                         & AP$^{bb}$     & AP$^{bb}_{50}$ & AP$^{bb}_{75}$ & AP$^{bb}$     & AP$^{bb}_{50}$ & AP$^{bb}_{75}$ & AP$^{mk}$     & AP$^{mk}_{50}$ & AP$^{mk}_{75}$ & AP$^{mk}$     & AP$^{mk}_{50}$ & AP$^{mk}_{75}$ \\ \hline
Supervised                                    & 38.9          & 59.6           & 42.0           & 40.6          & 61.3           & 44.4           & 35.4          & 56.5           & 38.1           & 36.8          & 58.1           & 39.5           \\ \hline
MoCo v2                                       & 39.2         & 59.9           & 42.7           & 41.5          & 62.2           & 45.3           & 35.7          & 56.8           & 38.1           & 37.5          & 59.1           & 40.1           \\ \hline
CLIM                                           &\textbf{39.5} &\textbf{60.0}  & \textbf{43.3}  & \textbf{41.8} & \textbf{62.3}  & \textbf{45.7}  & \textbf{35.8} & \textbf{57.0}  & \textbf{38.6}  & \textbf{37.7} & \textbf{59.4}  & \textbf{40.5}  \\ \hline
\end{tabular}
\end{center}
\vspace{-0.1cm}
\end{table}

\begin{table}[t!]
\vspace{-0.5cm}
\caption{Transfer learning on LVIS long-tailed instance segmentation (5 trials)}
\label{lvis-table}
\setlength{\abovecaptionskip}{0.3cm}
\setlength{\belowcaptionskip}{-0.1cm}
\renewcommand\arraystretch{1.1}

\begin{center}

        \begin{tabular}{l|ccc|ccc}
            \hline
            \multirow{2}{*}{Method} & \multicolumn{3}{c|}{Object Det} & \multicolumn{3}{c}{Instance Seg} \\ \cline{2-7}
                                     & AP$^{bb}$     & AP$^{bb}_{50}$ & AP$^{bb}_{75}$ & AP$^{bb}$   & AP$^{mk}_{50}$ & AP$^{mk}_{75}$        \\ \hline
            Supervised               & 24.1        & 39.4       & 25.0       & 24.2         & 37.8         & 25.1        \\ \hline
            MoCo v2                  & 25.1        & 40.4       & 26.1       & 25.3         & 38.4         & 27.0        \\ \hline
            CLIM & \textbf{25.5}        & \textbf{41.2}       & \textbf{26.7}       & \textbf{25.6}         & \textbf{39.5}         & \textbf{27.5}
            \\ \hline

\end{tabular}

\end{center}
\vspace{-0.4cm}
\end{table}
We also evaluate our feature representation on several downstream tasks, including object detection and instance segmentation, to evaluate the transferability of the learned features. For fair comparison, all experiments follow MoCo settings.

\textbf{PASCAL VOC Object Detection.}
Following the evaluation protocol in~\citep{he2020momentum}, we use Faster R-CNN~\citep{ren2015faster} with R50-C4 as backbone. We fine-tune all layers on the trainval set of VOC07+12 for 2$\times$ schedule and evaluate on the test set of VOC2007. We report the performances under the metric of AP50 and AP75. As shown in Table \ref{voc-table}, on PASCAL VOC, CLIM achieves $82.8\%$ and $64.5\%$ mAP under AP50 and AP75 metric, which is 1.4 points and 5.7 points higher than the fully supervised counterparts, and is slightly better than the results of MoCo v2.

\textbf{COCO Object Detection and Instance Segmentation.}
We also evaluate the representation learned on a large scale COCO dataset. Following \citep{he2020momentum}, we choose Mask R-CNN with FPN as backbone, and fine-tune all the layers on the train set and evaluate on the val set of COCO2017. In Table.\ref{coco-table}, we report results under both 1$\times$ and 2$\times$ schedules.  We show that CLIM consistently outperforms the supervised pretrained model and MoCo v2. Under 2X schedule, we achieve $41.8\%$ and $37.7\%$ detection and segmentation accuracies, respectively, which is 1.2 points and 1.1 points better than the supervised couterparts, and also slightly better than the highly optimized MoCo v2.

\textbf{LVIS Long Tailed Instance Segmentation.}
Different from VOC and COCO where the number of training samples is comparable, LVIS is a long-tailed dataset, which contains more than 1200 categories, among them some categories only have less than ten instances. The main challenge is to learn accurate few shot models for classes among the tail of the class distribution, for which little data is available. We evaluate our features on this long-tailed dataset to validate how the unsupervised representation boosts the performance. Similarly, we fine-tune the model (Mask R-CNN, R50-FPN) on the train set and evaluate on the val set of Lvis v0.5. Table.\ref{lvis-table} shows the result under 2$\times$ schedule. CLIM outperforms the supervised pretrained model by a large margin and slightly better than MoCo v2. We claim that it is mainly to the proposed data mixing data augmentation, which is able to learn generalized representations even with extremely few labeled data.

\subsection{Ablation Study} \label{ablation_part}
In this section, we present ablation studies to better understand how each component affects the performance. Unless specified, we train the model for 200 epochs over the ImageNet-1000 and report the top-1 classification accuracy under linear evaluation protocol.

\textbf{Positive Sample Selection.}
We first analyze the advantages of our proposed center-wise local sample selection strategy. The compared sample selection alternatives include:

$\bullet$ Random selection: Randomly select a sample from all unlabeled data.

$\bullet$ KNN selection: Use k-nearest neighbors to build the correlation map among samples, and randomly select a sample from the Top-$k$ ($k=10$) nearest neighbors as positive sample.

$\bullet$ K-means selection: Use k-means clustering algorithm to obtain $k$ cluster centers, and randomly select a sample from the corresponding cluster as positive sample.

$\bullet$ KNN $\cap$ K-means selection: Use K-means clustering algorithm to obtain $k$ cluster centers, and randomly select nearest neighbor within the cluster as positive sample.

The results are shown in the second column of Table.\ref{pos-table}. In order the inspect the influence of sample selection, we do not conduct cutmix augmentation, and these positive samples are simply pulled via a standard contrastive loss. Noted that, we test many hyper-parameter settings for each sample selection strategy, and choose the best settings for them. It can be shown that comparing with the MoCo baseline, both KNN and cluster-based sample selection boost the performance, while our proposed center-wise selection surpasses these two methods by $1\%$ and $1.3\%$, respectively. 

\begin{figure}[t]
\begin{minipage}[t!]{\textwidth}
 \begin{minipage}[t!]{0.45\textwidth}
  \centering
        \makeatletter\def\@captype{table}\renewcommand\arraystretch{1.0}\renewcommand\tabcolsep{3pt}\setlength{\abovecaptionskip}{0.25cm}\setlength{\belowcaptionskip}{0.2cm}\makeatother\caption{Impact of different sample selection}\label{pos-table}
        \begin{tabular}{lcc}
            \hline
            \multirow{2}{*}{Strategy}& \multicolumn{2}{c}{ Accuracy ($\%$)}
             \\ \cline{2-3}
                         &   no mixing      &  +cutmix   \\ \hline
            MoCo v2             &    67.5    &    -   \\
            Random             &    62.3    &    67.1   \\
            KNN        &    68.3    &    69.5   \\
            K-means      &    68.0    &    69.2 \\
            KNN $\cap$ K-means        &    68.5    &    69.6   \\ 
            \hline
            Center-wise &             \textbf{69.3} &   \textbf{70.1}  \\

            \hline

        \end{tabular}

  \end{minipage}
  \hspace{0.2cm}
  \begin{minipage}[t!]{0.5\textwidth}
   \centering
        \makeatletter\def\@captype{table}\renewcommand\arraystretch{1.29}\renewcommand\tabcolsep{3pt}\setlength{\abovecaptionskip}{0.25cm}\setlength{\belowcaptionskip}{0.25cm}\makeatother\caption{Impact of different multiple resolutions }\label{mr-table}
        \begin{tabular}{lcr}
            \hline
            Method                           & Resolution                          & Accuracy ($\%$) \\ \hline
            Multi-Crop                        &  2$\times$224 + 2$\times$96         & 69.7           \\
            \hline
            \multirow{4}{*}{Multi-Reso} &  $r,{r}' \in \{224, 96\}$               &   70.4                \\
                                     &  $r,{r}' \in \{224, 128\}$               &   71.7  \\
                                              & $r,{r}' \in \{224, 160\}$      & \textbf{72.3}   \\
                                              &  $r,{r}' \in \{224, 224\}$    & 71.4            \\ \hline
        \end{tabular}

   \end{minipage}

\end{minipage}
\vspace{-0.3cm}
\end{figure}

\textbf{Data Mixing Augmentation.}
Data mixing helps to expand the neighborhood space of the target sample, and acts as smoothing regularization for the prediction. As shown in the third column of Table.\ref{pos-table}, cutmix augmentation consistently improve the performance, comparing with directly pulling similar samples in contrastive loss, and achieve $70.1\%$ accuracy with only 200 training epochs. Notably, with randomly selected positive samples, cutmix operation even obtains $67.1\%$ accuracy, slightly lower than the MoCo baseline, while significantly better than no mixing with only $62.3\%$ accuracy. This can be attributed to the smoothing regularization of cutmix, which is able to alleviate the effect of noisy samples and update model in a more robust way.

%
%

\textbf{Multiple Resolution.}
Based on CLIM, we further add multi-resolution data augmentation to validate its effectiveness. The results of introducing different resolutions are shown in Table.\ref{mr-table}. Using multiple resolutions setting with $r,{r}' \in \{224, 160\}$, our method achieves an accuracy of $72.3\%$ with only 200 epochs, which surpasses the baseline of MoCo by $4.8\%$, and even much better than the results of MoCo with 800 epochs ($71.1\%$).
We also compare our multi-resolution augmentation with multi-crop augmentation proposed in \citep{caron2020unsupervised}. $2\times 224 + 2\times96$ denotes using two $224 \times 224$ crops with crop-scale $\sigma \sim U(0.2,1.0)$ and two $96 \times 96$ crops with $\sigma \sim U(0.05,0.14)$, referring to \citep{caron2020unsupervised}.  We find that multi-crop slightly deteriorates the performance of CLIM ($70.1\%$ versus $69.7\%$), partially because data mixing behaves like image cropping augmentation, and shares similarity with multi-crop strategy.

\textbf{More Ablation Studies}. 
We provide more ablation studies for reference in the Appendix. Detailed comparisons include 1) Number of clusters $m$ and $k$ in KNN, 2) Hyperparameters $\alpha$ in Cutmix , 3) Different choices of mixing methods and  4) Ablation study on longer training schedule.

\section{Conclusion}
In this work, we proposed CLIM data augmentation. Center-wise positive sample selection considers both local similarity and global aggregation property. In such way, similar samples are progressively aggregated to a series of predefined clusters, while not breaking the local similarity. Data mixing augmentation acts as a smoothing regularization for contrastive loss between neighborhood space. Furthermore, we present a simple but effective multi-resolution augmentation, which explicitly model scale invariance to further improve the representation. Experiments evaluated on several unsupervised benchmarks demonstrate the effectiveness of our method.

\bibliography{egbib}
\end{document}